\documentclass[journal]{vgtc}                
\ifpdf
  \pdfoutput=1\relax                   
  \usepackage{graphicx}                
  \DeclareGraphicsExtensions{.pdf,.png,.jpg,.jpeg} 
\else
  \usepackage{graphicx}                
  \DeclareGraphicsExtensions{.eps}     
\fi%

\graphicspath{{figures/}{pictures/}{images/}{./}} 

\usepackage{microtype}                 
\PassOptionsToPackage{warn}{textcomp}  
\usepackage{textcomp}                  
\usepackage{mathptmx}                  
\usepackage{times}                     
\usepackage{cite}                      
\usepackage{tabu}                      
\usepackage{booktabs}                  

\usepackage{enumerate}
\usepackage{graphicx}
\usepackage{amssymb}
\usepackage{amsmath}
\DeclareMathOperator*{\argmax}{argmax} 
\usepackage{algorithm}
\usepackage[noend]{algpseudocode}
\usepackage{commath}
\usepackage{caption}
\usepackage[caption=false, font=footnotesize]{subfig}
\usepackage{setspace}
\usepackage{enumitem}
\usepackage{xcolor}
\definecolor{mybrown}{RGB}{106, 74, 60}
\definecolor{myred}{RGB}{204, 50, 63}
\definecolor{mypurple}{HTML}{5F259E}
\definecolor{mygreen}{HTML}{67B826}
\onlineid{1145}

\vgtccategory{Research}
\vgtcpapertype{System}

\title{Interactive Visualization for Explaining Value-level \\Feature Attribution in Event Prediction with RNNs}

\author{Chuan Wang, Xumeng Wang, and Kwan-Liu Ma}
\authorfooter{
\item
 Chuan Wang is with University of California, Davis. E-mail: nauhcy@gmail.com.
\item
 Xumeng Wang is with Zhejiang University. E-mail: wangxumeng@zju.edu.cn.
\item
 Kwan-Liu Ma is with University of California. E-mail: ma@cs.ucdavis.edu.
}

\shortauthortitle{C.W \MakeLowercase{\textit{et al.}}: ViSFA}

\abstract{Deep Recurrent Neural Networks (RNN) is increasingly used in decision-making with temporal event sequences. However, understanding how RNN models produce final predictions remains a major challenge. Existing work on interpreting RNN models for sequence predictions often focuses on explaining predictions for individual data instances (e.g., patients or students). Because state-of-the-art RNN models are formed with millions of parameters optimized over millions of instances, explaining predictions for single data instances can easily miss a bigger picture. Besides, RNN models often use multi-hot encoding to represent the presence/absence of features, where the interpretability of feature attribution with numeric values is missing. We present ViSFA, an interactive system that visually summarizes feature attribution over time for different feature values.
ViSFA scales to large data such as the MIMIC dataset containing the electronic health records of 1.2 million high-dimensional temporal events. We demonstrate that ViSFA can help us reason RNN prediction and uncover insights from data by distilling complex attribution into compact and easy-to-interpret visualizations.%
} 


\vgtcinsertpkg

\begin{document}



\maketitle


\section{Introduction}
Deep Recurrent Neural Networks (RNN) is pervasively used in reasoning and decision-making tasks for sequential data analysis. Due to its remarkable performance and broadly applicable feature, RNNs have helped solve problems in domains from fundamental research such as natural language processing (NLP) \cite{Young2018} and video analysis to domain-specific research such as electronic health records (EHR) analysis \cite{Choi2016}, customer behavior analysis \cite{WangONM18}, and stock prediction \cite{Bao2017}. 

Despite RNNs' popularity and remarkable performance, end-users' demand on model \textit{trust}, \textit{fairness} and \textit{reliability} makes its limited use in many critical real-world decision-making schemes.
One would be critical of a reliability-oriented paper that only cites accuracy statistics \cite{DoshiKim2017Interpretability}. Consciously collected data can be easily biased, and models built with such data can be unreliable. For example, Ribeiro et al. \cite{Ribeiro2016} show a case where the snow backgrounds in the training images, instead of real morphological features, distinguishes ``huskies'' from ``wolves''. Such problems also happen to sequential data analysis. 
It will be critical and causing fatal issues if similar problems happened to applications such as health care, autonomous vehicles, or legal matters.
How to make sure a model's decision is not based on biased facts? How do we provide trusted prediction systems that humans are confident to use? How do we guarantee the decisions are not discriminated against by a special group? 
It's urgently demanded to solve these problems so we can use machine learning models transparently and safely.

To address this subject, existing work has focused on visualizing RNN predictions for domain-specific tasks such as NLP \cite{Ming2017, Strobelt2016LSTMVisAT} by revealing RNNs' inner mechanisms.
However, RNN models have \textbf{a broader usage in various scientific research fields} such as DNA sequence analysis \cite{Quang2016}, electronic health records (EHR) analysis \cite{Choi2016}, customer purchase intent analysis \cite{Sheil2018PredictingPI}, stock prediction \cite{Bao2017}.
Explanations in domain-specific manners can not fulfill the requirements in diverse scenarios. For example, word-to-vector embeddings in the NLP domain are fundamentally different from one-hot or multi-hot embeddings in most applications where each embedding vector dimension has physical meaning. Owing to this difference, the visual analysis for the NLP domain can be distinctive among general visual analysis with RNN predictions.
Besides, interpreting inner-model behavior can be fragmentary in real practice and can potentially cause catastrophic harm to society \cite{Rudin2018}. 
Furthermore, it's important for humans to understand and trust a model's predictions by knowing how multi-dimensional features and their temporal changes \textbf{in entire prediction classes} contribute to prediction. Current visual interpretations often focus on explaining predictions for individual instances \cite{KwonCKCKKSC19, Olah2017}. Because deep learning models are computed based on a populous distribution of instances, explanations for individual instance prediction can easily miss a bigger picture of valuable insights learned by a model.  
Last, \textbf{value-level feature attribution} is not previously addressed for model interpretation, yet explaining what values in a feature contribute to a particular class would be more meaningful than simply illustrating whether a feature has a high contribution. For example, knowing the importance of feature ``customer visits''  would provide less guidance than telling how many visits would be effective in preserving customers. 
Therefore, it's often desirable to explain models by visualizing \textbf{temporal feature value attribution for the entire prediction classes in a domain-independent manner}. 

Our work is inspired by these unfulfilled requirements and attempts to resolve the following challenges. 
\textit{Data complexity}: datasets for RNN modeling can be seen as tensors that composed of time, multi-dimensional feature, and instance. In the settings of RNN modeling, each instance is a sequence and is labeled with a class. For example, in the patient mortality analysis, a patient's medical temporal history is associated with a mortality label dead/alive. At each time-step, a temporal event is composed of multi-dimensional features representing values from medications, lab test results, etc. Despite the complexity in data rank and dimension, analyzing such practical data often desires the handling of specialties such as sparsity and noisiness.  
\textit{The dilemma -- generality or visual complexity}: 
Complex visualizations systems are often tailor-made for specific data characteristics or analytic tasks, and high generality requirements often limit possible visualization designs. We aim to handle compound computations at the backend in exchange of easy-to-comprehend visualization designs for a model-agnostic straightforward interpretation of data insights.
We introduce ViSFA, a visual analytics system that summarizes feature attribution in temporal sequences with RNNs. Our contributions are:

\begin{itemize}[noitemsep,topsep=0pt,parsep=0pt,partopsep=0pt,leftmargin=8pt]
    \item A scalable and domain-independent visual analytics approach that summarizes feature attribution in entire prediction classes. We designed a series of modular algorithms based on stratified sampling and gap statistic clustering, which facilitate a fair comparison of contributing patterns in temporal value change between classes. 
    \item An \textbf{interactive} visualization system for users to remove irrelevant noises and discover major contributing sequential patterns. 
\end{itemize}

\section{Related Work}
A deep neural network is typically trained with a large number of instances that each is annotated with a class label. A well-trained deep learning model can learn useful knowledge from the instances and form a complex network of neurons that transform the instances to a probability for class prediction. However, how deep neural networks learn knowledge and achieve outperform predictions are remained unclear. 
Researchers from both the Machine Learning and Data Visualization field have investigated the interpretability of deep learning models. However, there is not a single interpretation method that can be applied to any deep learning models \cite{DoshiKim2017Interpretability}. 

\subsection{RNN Model interpretion}
Currently, there are two major approaches in terms of whether RNN models are extended for interpreting deep learning models. 

To interpret the dynamics of deep neural networks, a few studies apply visual approach to explain convolutional neural networks \cite{Rauber2017, Kahng2017ActiVis, Simonyan2013DeepCN, Paulo2016, Bilal2018}. For understanding RNNs, model-generating factors, like hidden states, need to be explored. LSTMVis  \cite{Strobelt2016LSTMVisAT} visualizes the hidden state dynamics of RNNs by a parallel coordinates plot. To match similar top patterns, LSTMVis allows users to filter the input range and check hidden state vectors from heatmap matrices. For natural language processing tasks, Ming et al.~\cite{Ming2017} also employ a matrix design. Combining hidden state clusters and word clusters, Ming et al.~\cite{Ming2017} design a co-clustering layout, which links cluster matrices and word clouds.

In addition, extending RNNs by extra structures, like neural attention mechanisms, contributes to easier interpretation. The attention mechanism has become popular recently because the added attention layers allow the interpretation of a particular aspect of the input \cite{Bahdanau2014, Xu2015ShowAttendnTell, Hermann2015TeachingMT}. Besides, the attention mechanism is proved to be able to improve the performance of deep learning models. 
Hermann et al. \cite{Hermann2015TeachingMT} develop an attention-based method for deep neural nets for comprehending documents. Due to easier interpretation for end-users, this group of research is often applied to real-world analysis in a variety of domains. However, only a few works focus on the visual interpretation of deep learning with an attention mechanism. RetainVis \cite{KwonCKCKKSC19} is a rare example that is an interpretable and interactive visual analytics tool for EHR data.


\subsection{Additive Feature Attribution Methods}
The highest accuracy for large modern datasets is often achieved by complex models that even experts struggle to interpret. To simplify model understanding, a branch of research uses additive feature attribution methods that consider deep learning models as black boxes and explain simpler explanation models as approximations of the original model. 
Model-specific approximation such as DeepLIFT \cite{Shrikumar2017PMLR, Shrikumar2016} compares the activation of each neuron to its ``reference activation'' and assigns contribution scores according to the difference. 
There are also model-agnostic methods such as LIME \cite{Ribeiro2016} and SHAP \cite{Lundberg2017}. LIME interprets individual model predictions based on locally approximating the model around a given prediction.
SHAP use Shapley values as a measure of feature variable contribution towards the prediction of the output of the model.
Simpler explanations improves computational performance for interpretation. 
Similar in spirit, Manifold \cite{ZhangWMLE2019Manifold} provides a visual analysis framework to support interpretation, debugging, and comparison of machine learning models. These approaches differ from our work in that they approximate RNN models with simpler explanation models, whereas we attempt to interpret the behavior of original models.

\subsection{RNN Application fields} 
RNN becomes popular for its high performance in the linguistic domain. Most well-known works are from the NLP field, such as machine translation and sentiment analysis \cite{KarpathyJL15}. Due to the popularity of NLP tasks, the majority of RNN interpretation work has made efforts to explain RNN models under the NLP background, such as \cite{Ding2017, Li2016VisNLP, Ming2017, Strobelt2016LSTMVisAT}. 
However, the visual analysis of NLP often requires special visualization techniques because language composed by individual words is a unique type of data. 
The architecture of RNN is distinctive because connections between nodes form a directed graph along a temporal sequence. Therefore, RNN can be used to a broad range of sequence analysis applications, such as DNA sequence analysis \cite{Quang2016},  Electronic Health Records (EHR) analysis \cite{Choi2016}, customer purchase intent analysis \cite{Sheil2018PredictingPI}, stock predictions \cite{Bao2017}, and so on. 
As discussed in the introduction, the interpretability is critical to these application domains regarding model fairness, reliability, and trust. 
However, not much attention has addressed to the visual interpretation of RNN from a more generic perspective except for ProSeNet \cite{Xu2019kdd}, which proposes an interpretable and steerable deep sequence model.  

This paper attempts to bridge this gap and propose a visual analysis method that can be applied to broader scenarios of analyses for summarizing contributing sequence patterns in predictions. Because end-users' understanding is an urgent desideratum, this work focuses on visualizing distilled patterns that directly map to the original data, based on attention mechanism enhanced RNN model for its easy-to-perceptible superiority. Besides, to our best knowledge, there is no work explaining how RNN models correlate contributing numeric/categorical values with the prediction except AttentionHeatmap \cite{WangONM18}. This work extends AttentionHeatmap in the analysis flow and visualization designs. 

\section{Design Considerations}
There are a few observations of RNN models before visualizing feature attribution with them \cite{Ming2017, Strobelt2016LSTMVisAT, WangONM18}.
First, if an RNN model is successfully trained, the instances within a prediction class in the dataset are expected to share some common characteristics which can be captured by the RNN model. In other words, if the instances do not share any common pattern within any prediction class, it's impractical to learn a convergent or high-performance RNN model.
Second, characteristics from different prediction classes are different. If the commonalities from one class cannot distinguish itself from another, the model training cannot succeed. Third, the attention weight of an event reflects the importance of the event in making such distinction. However, because state-of-the-art RNN models trained with real-world datasets can hardly achieve 100\% accuracy, the learned attention weights are often not completely accurate. For example, LSTMVis \cite{Strobelt2016LSTMVisAT} notices interpretable patterns but also significant noise when studying RNN models. Likewise, AttentionHeatmap \cite{WangONM18} reveals that the filtered events contain noise that does not follow the major pattern no matter what attention range is selected by the user. 



Based on the above observations, we derive the following design goals (\textbf{DG}s):

    \textbf{DG1: Facilitate the attribution analysis for tensors that are composed of dimensions including time, instance, feature, and feature value.} Given multi-dimensional tensors, how do we synthesize meaningful visualizations for interpreting a model's prediction? Knowing how entire classes share common patterns is not enough because these patterns not necessarily contribute to the model formation, a.k.a, distinguishing different classes. Therefore, the system should help to distill complex data to find the contributing subset and visualize the patterns in the subset. 
    
    \textbf{DG2: Highlight major patterns across the instances within each prediction class.} 
    As mentioned earlier, we notices significant noise when studying deep learning models. The learned attention weights are noisy too. The visualization should remove or minimize the influence of noises and highlight major patterns in data. 
    Besides, the visualization should highlight common patterns shared among the instances in a prediction class. Those common patterns are the keys for users to find insights from each class. 
    
    \textbf{DG3: Contrast differences between prediction classes.} The visualization design should facilitate easy and fair comparison between different classes. For example, visual comparison based on imbalanced class sizes can suffer from inequity if the visualization results are affected by class sizes. 
    The visualization design should guarantee the visualized pattern is a true reflection of its belonging class instead of the influence of class size.
    
    \textbf{DG4: Be able to scale for large datasets.}
    Because predictions based on state-of-the-art models are formed with millions of weights optimized over millions of data instances, explaining predictions for single data instances can miss a bigger picture oftentimes. Understanding how entire classes contribute to a model is important for trusting a model's prediction and deciphering what a model has learned. Therefore, the design should build entire class representations regardless of the class size. 
    
    \textbf{DG5: Be generic for different applications.} RNN becomes widely used across different domains because of its generality. Even the vanilla RNN models can be adaptive for multiple disciplines such as finance stock price forecasting \cite{Bao2017} and  
    customer analysis predicting purchasing intent \cite{Sheil2018PredictingPI}. 
    It's challenging but meaningful to build a domain-independent visualization system. For users from different domains, we should provide easy-to-interpret interaction and visualization designs. 

\section{RNN Model}

This approach leverages LSTM attention neural networks. The attention model computes a value that represents the importance of a particular temporal event at every time-step for all instances. Existing approaches build end-to-end models for RNN interpretation such as \cite{KarpathyJL15, Choi2016}. These methods are not sufficiently developed for many real-world applications. They often use multi-hot vector feature encoding where 0/1 are used to represent the appear/absence of a feature but without detailed information. For example, in the EHR data analysis, each temporal event is a patient visit. During a patient visit, multi-hot encoding records whether a treatment is performed or a type of medicine is applied. However, either using a treatment/medication or not is more of a standard process. It's  more meaningful to study how much treatment or the dose of medication can be more effective to a patient group. 

Our work solves the problem by using numeric feature encoding so that the analysis result can help understand the importance of different feature values. 
Numerical encoding contains higher granularity of information compared to multi-hot encoding. For example, the information `a patient's taking antibiotics every day’ is less informative compared to know a patient’s antibiotics doses every day. 
It is more difficult to tune a model with data of high granularity due to its complexity.
However, by including actual values in the learning process, the analysis can be more practical for studies. In the above example of EHR data analysis, our approach keeps track of the contribution of each feature value. The following contribution analysis 
more likely provides insights that can help doctors make future decisions such as what treatment to apply to patients.

\section{ViSFA}
AttentionHeatmap \cite{WangONM18} presents a visual analytics interface for RNN feature attribution analysis, which is composed of the matrix grid view for visualizing time-folded feature attribution, and the k-partite graph view for displaying attribution change over time. 
The interface of AttentionHeatmap is designed for user groups such as RNN model trainers and data scientists. For ultimate end-users like domain analysts who have no machine learning knowledge, we create ViSFA to help them focus on major patterns distilled from data. For time-folded feature attribution analysis, ViSFA improves AttentionHeatmap by integrating a ranking algorithm that sort features in contribution descending order. ViSFA improves the visual analytics procedures by providing comparable visual summaries of temporal feature attribution and lets end-users interactively removing noise (\textbf{DG1}). 
We introduce more details in the following sections. 

\subsection{Data \& Control Flow}
We pre-process data so that the time-steps for each instance are aligned reasonably. 
Each instance $x_{i}$ in the tensor space $\mathbb{R}_{I}^{T \times F}$ is a temporal sequence $\{x_{i}^{1}, .., x_{i}^{t-1}, x_{i}^{t}, x_{i}^{t+1}, ..., x_{i}^{T}\}$ that is of length $T$ and labeled with a class $y_{i}$ in the training dataset $D = \{x_{i}, y_{i}\}$, where $t \in \{1, T\}$ and $y \in \{1, L\}$ is one of $L$ categorical labels. 
$f \in \{1, F\}$ is one of $F$-dimensional features whose value range is $[v_{f_{min}}, v_{f_{max}}]$, and each event $x_{i}^{t}$ can be further expanded to a multidimensional expression $\{x_{i}^{t,1}, .., x_{i}^{t, f-1}, x_{i}^{t, f}, x_{i}^{t, f+1}, ... , x_{i}^{t, F} \}$ where $x_{i}^{t,f} \in [v_{f_{min}}, v_{f_{max}}]$.  Therefore, the research problem becomes calculating contribution scores $s_{f}$ for each feature, and visualizing the temporal change in the values of highly contributing features. But how to calculate the contribution of each feature? 

Fortunately, the attention mechanism for RNN models is designed to solve the problem. 
The RNN model training process is using the dataset $D$ to form a complex model $\hat{y}_{i} = g(x_{i})$ where $g$ transforms input data $\mathbf{x}$ to $\mathbf{y}$ using millions of parameters. During the training process, parameters were decided by minimizing a chosen loss function $\sum_{t=1}^{T} \mathbb{L}_{t} (y_{i}^{t}, \hat{y}_{i}^{t})$. 
The attention mechanism plugs an attention network to original RNN model and learns a set of parameters $\{a_{i}^{1}, .., a_{i}^{t-1}, a_{i}^{t}, a_{i}^{t+1}, ..., a_{i}^{T}\}$ where $a_{i}^{t} \in [0,1]$ represents the importance of each event in the corresponding instance's temporal sequence. The two-level attention model further computes a set of parameter $\{a_{i}^{t,1}, .., a_{i}^{t, f-1}, a_{i}^{t, f}, a_{i}^{t, f+1}, ... , a_{i}^{t, F} \}$ at each temporal event $a_{i}^{t}$ for each corresponding feature.


\begin{figure*}
    \centering
    \includegraphics[width=\textwidth]{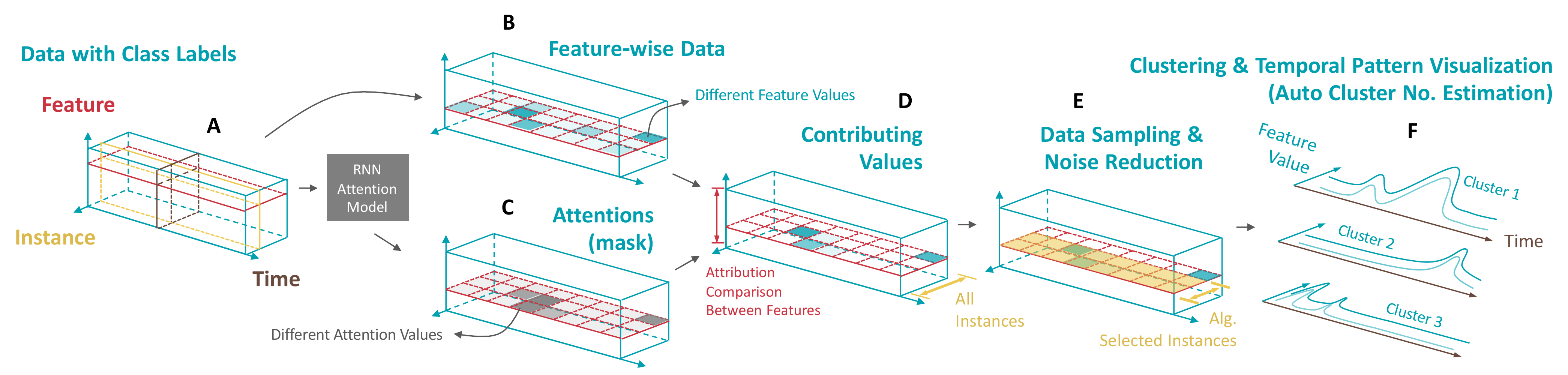}
    \caption{Workflow of ViSFA.}
    \vspace{-10px}
    \label{fig:flow}
\end{figure*}

Figure \ref{fig:flow} illustrates the high-level workflow of ViSFA. 
The visual summary module is composed of part1) the time-folded summary module and part2) the module for visually summarizing contributing temporal patterns. We applied the matrix grid view in AttentionHeatmap~\cite{WangONM18} in the time-folded summary module. Our work enhances the matrix-grid view with a ranking algorithm that comprehensively computes the contribution of each feature. ViSFA focuses on the temporal pattern analysis and Figure  \ref{fig:flow} illustrates part2. 

First, the tensor of multidimensional temporal sequences is feed into an RNN model to train a classifier that can predict the class for all sequences. The RNN model must be validated and have an user-verified prediction performance. The attention mechanism then outputs the attention sequence $\mathbf{a}^{T}$. Thereafter, two sets of data are the input of the visualization module: the original data containing instances and their labels $D = \{x_{i}, y_{i}\}$ (B) and the attention sequences $\mathbf{a}^{T}$ (C) that are associated with each temporal sequence and reflects the importance of each temporal event in the sequences. The way C works is like a mask applied to B. A user can filter the original temporal events with an attention range of interest (AOI).
Specifically, after user selecting AOI, the feature values whose corresponding attention values are outside the AOI becomes ``NULL'' in the vector representation (D). In deep learning, empty events are usually padded with zeros for easy matrix computation. However, in our visual analysis step, padding zeros for the unselected events can be ambiguous because zero values can both represent uninterested events and contributing values. Therefore, the feature $f$'s value range becomes $\{\emptyset \cup [v_{f_{min}}, v_{f_{max}}]\}$. Particularly, zero values of feature can be important if their contribution is high.

ViSFA then transforms the filtered data with a sequence of operations such as data sampling and noise reduction (E), as shown in Figure \ref{fig:flow}. To show the common temporal patterns distilled by a series of algorithms, ViSFA clusters the temporal sequences and automatically estimates the number of clusters (F). 
To fulfill the design goals, we tested several operational and visualization design alternatives and resolved with a series of algorithms that keep users in the loop. 
We explain the details in the following sections.

\subsection{Ranking Feature Contribution}
Two factors are taken into consideration when determining the overall contribution score $s_{f}$ for feature $f$: the average frequency of all feature values $C$ and the variance $V$. If the values of $f$ are divided into $M$ bins $b_{m}^{f}$ ($m = [1,M]$), the definition of $s_{f}$ is

\vspace*{-0.09in}
\begin{equation}
s_{f} = C(x^f) * V(x^f) 
\end{equation}
\vspace*{-0.09in}
where 

\vspace*{-0.08in}
\begin{equation}
C(x^f) = \frac{\sum_{m=1}^{M} |b_m^f|_c}{M}
\end{equation}
\vspace*{-0.08in}

\noindent
and $V(x^f)$ represents the variance of array $x^f$. $|x|_{c}$ represents the cardinality of $x$.
Term $C$ determines whether contributing feature values between classes are profoundly different on average.
Term $V$ measures how different the bin values contribute to classification. Therefore the score $s_{f}$ combines the contribution of feature values and their variations. Because $C>0$ and $V>0$, we have $s_{f} > 0$. 

AttributionHeatmap summarize time-folded feature attributions with matrix grids \cite{WangONM18}. 
We extend the matrix grid visualization with a ranking algorithm so that the feature of the highest score is on the top/left and the lowest score on the bottom/right.
With the ranking function, users can instantly locate features of interest (FOIs) instead of searching for FOIs by visually comparing matrices in the matrix grid. Users can also interactively select the number of features to remove the features of low contributions. 

After users located FOIs, a primary contribution of ViSFA is helping users to compare the contributing sequential patterns between different classes. There are challenges in such visual comparisons, such as different instance sizes in two classes and significant noises in computed feature attribution. We design a series of procedures to solve these problems. As shown in Figure \ref{fig:flow}, we present these procedures D, E and F in the following sections.
We provide a summary of these procedures in Algorithm \ref{algorithm}.

\subsection{Between-class Comparison -- Stratified Sampling} 
Fair comparisons are prerequisites (\textbf{DG3}) when comparing the contributing sequences between entire prediction classes. However, sequences from two prediction classes can be fundamentally different in size, distribution, and contributing values. Visually comparing them by simply aggregating the sequences within each class can be misleading. For instance, a between-class comparison showing the aggregated values for two imbalanced instance sizes can lead to ambiguous because it can be either the values or the instance sizes that causing an overall difference.
Besides, the analysis should be scale to summarize temporal patterns from large datasets (\textbf{DG4}). Even if the scale of a training dataset reaches or larger than millions, the analysis should still be able to visualize patterns for different classes in the dataset. For a large dataset, a particular scale of aggregation or summarization is essential. Otherwise, the visualization can suffer from a limited canvas size if using a juxtaposed design or a visual blocking if adopting an overlapped design.
Additionally, the summary of contributing sequences should be a true reflection of its belonging class.

To meet these requirements, we propose to use a method using stratified sampling after testing a few alternatives, such as random sampling. Stratified sampling is one of the probability sampling methods that sample equal size instances from all classes and sampling the most representative instances from each class. In statistics, stratification is the process of dividing members of the population into homogeneous subgroups before sampling. 
We choose to use stratified sampling to sample instances from data's subpopulations because sampled instances cover all possible subpopulations and is a substantive reflection of the original data population. 

\begin{figure}[!ht]
     \centering
     \subfloat[\label{fig:gapstatistic1} Example data set that homogeneously form two clusters and an outlier (17).]{%
         \centering
         \includegraphics[width=\columnwidth]{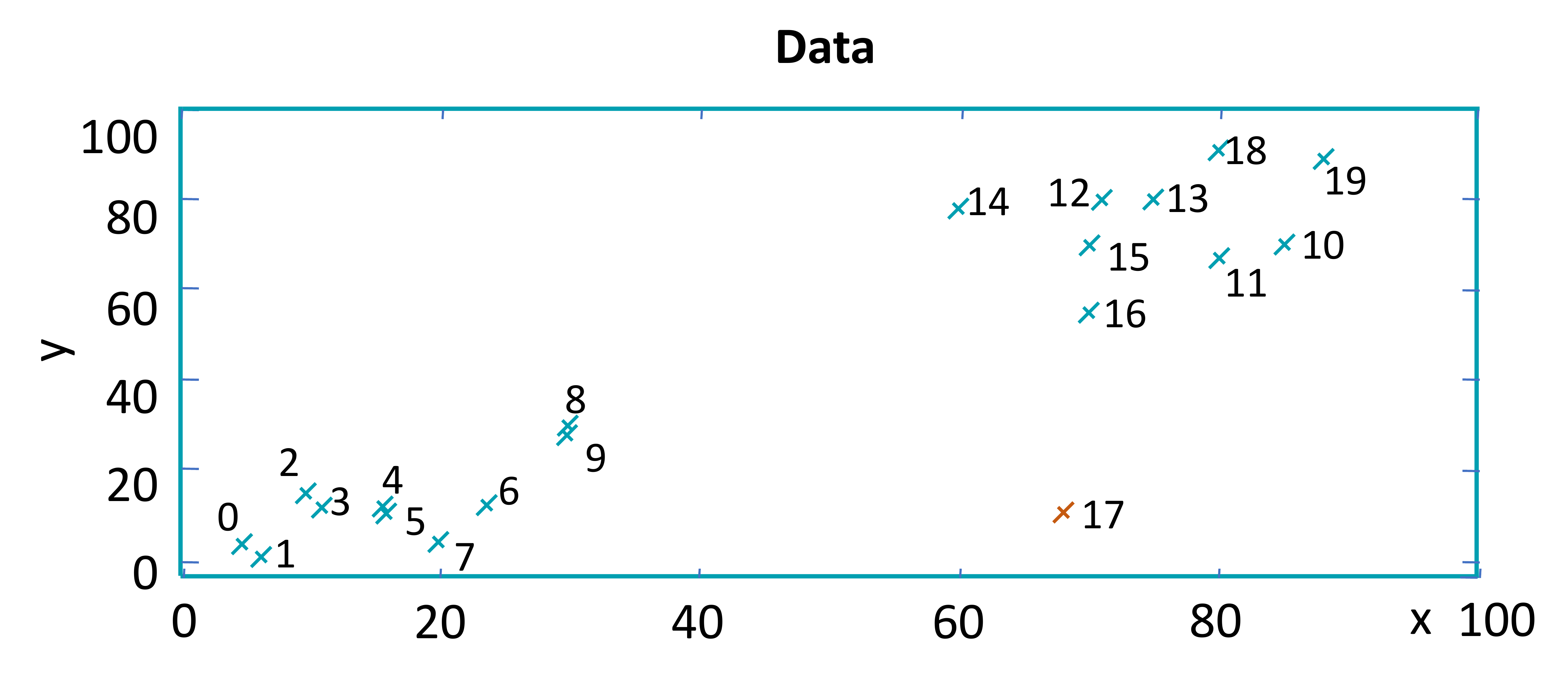}
     }
     \vspace{-8px}
     \hfill
     \subfloat[\label{fig:gapstatistic2}The dendrogram iteration vs. distance curve and the elbow point (blue dot).]{%
         \centering
         \includegraphics[width=0.98\columnwidth]{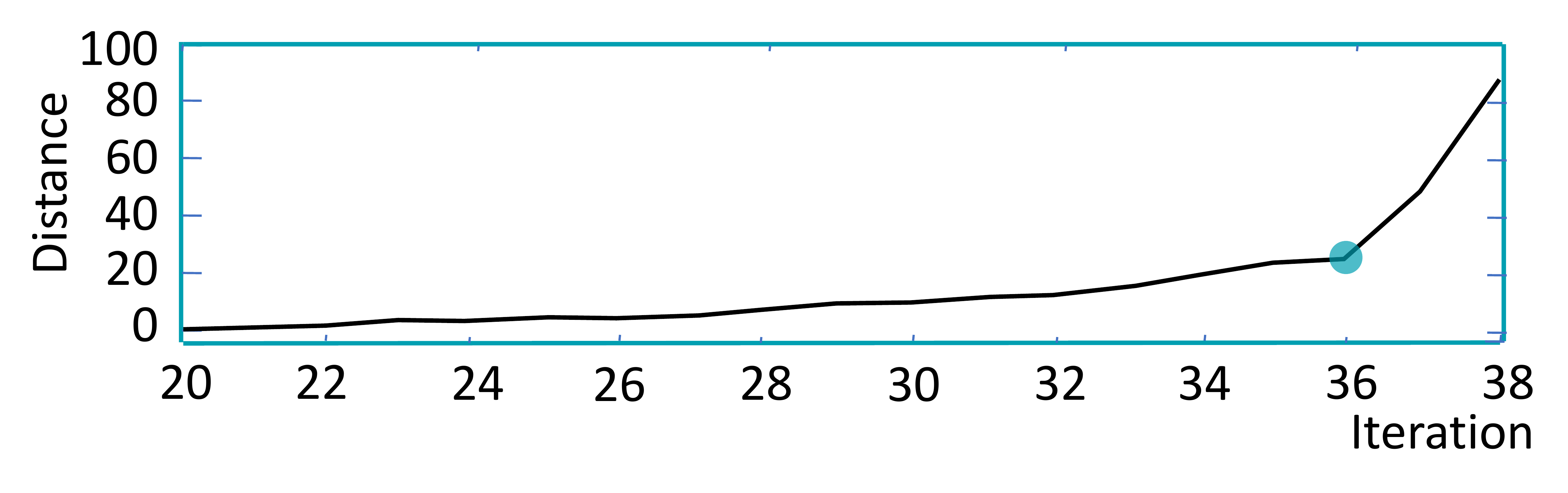}
     }
     \hfill
     \vspace{-8px}
     \subfloat[\label{fig:gapstatistic3}A dendrogram illustration of HAC-based outlier removal. Left: original data, locating outlier 17 with dendrogram cutting. Right: new dendrogram forms after removing outlier. Numbers represent node IDs in sequential order.]{%
         \centering
         \includegraphics[width=\columnwidth]{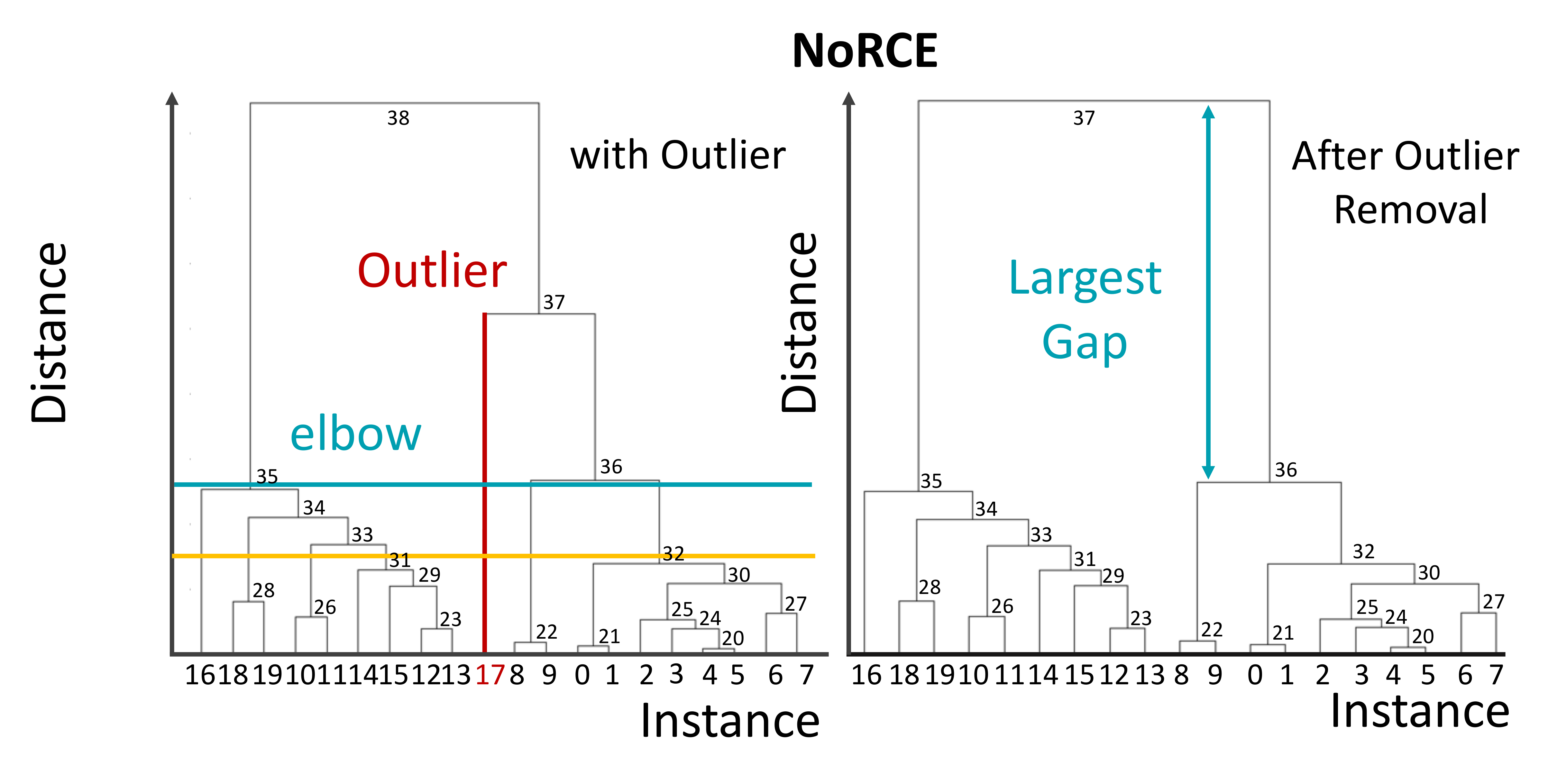}
     }
     \hfill
     \vspace{-8px}
     \subfloat[\label{fig:gapstatistic4}Optimal cluster size (dots) comparison \textcolor{myred}{before} and \textcolor{mybrown}{after} removing outlier.]{%
         \centering
         \includegraphics[width=\columnwidth]{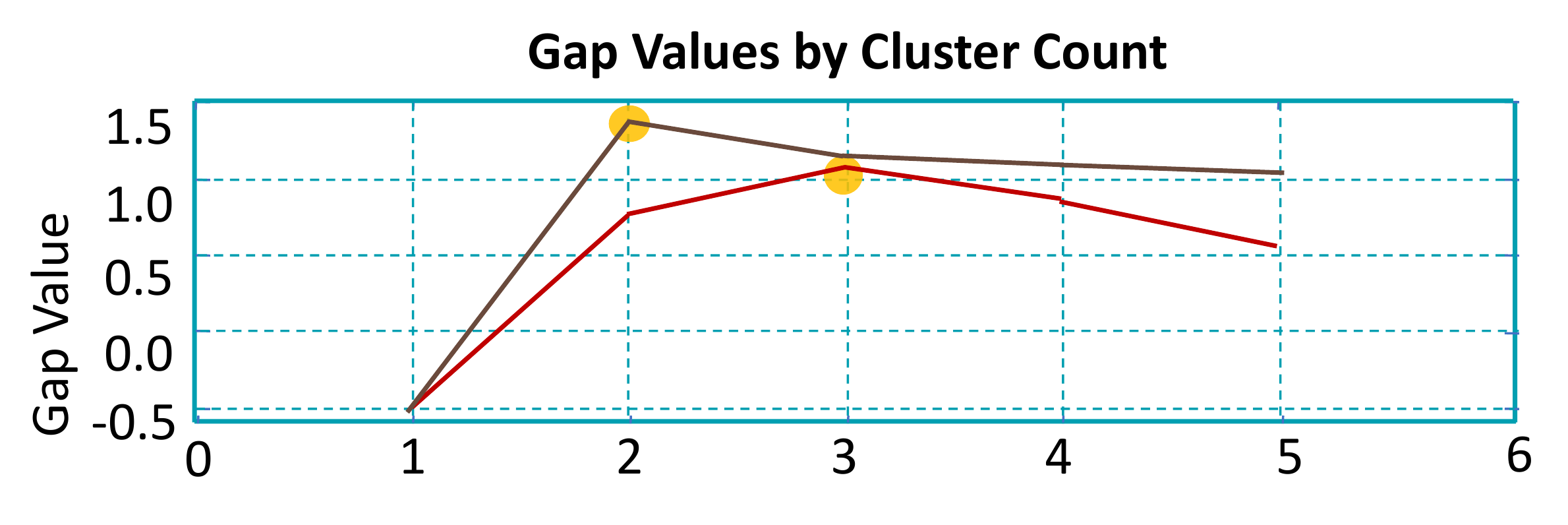}
     }
    
    \caption{Illustration of NoRCE in with a simple example. }
    \vspace{-5px}
    \label{fig:gapstatistic}
\end{figure}

However, it's non-trivial to find homogeneous subpopulations from a set of sequences $\mathbf{x}_{t}$ belonging to a particular class, because these sequences are often high-dimensional (in time steps) and noisy. Determining the number of clusters $S$ is difficult for unknown data distribution. Fortunately, the parameter $S$ is usually large for sampling tasks. For an example of our experiments, $S$ is on a scale of 3K for a balanced training dataset that each class contains 10K instances. That is, around $30\%$ of the instance population is sampled. We leverage the Hierarchical Agglomerative Clustering (HAC) that gradually calculates the increased distances between instances and newly formed dendrogram nodes in a bottom-up direction. The HAC iteration stops when the node size reaches $S$ to save computation time. As illustrated in a native example in Figure \ref{fig:gapstatistic4} left, the iteration stops at the cut location (yellow line) where only nodes below the cut are computed. The algorithm then randomly sample an instance from each sub-cluster to form $S$ samples in total. 
In Algorithm \ref{algorithm}, the sampling procedure includes a computational efficient HAC algorithm in lines \ref{lst:line:stratifiedsamplingHACBegin}-\ref{lst:line:stratifiedsamplingHACEnd}, except that line \ref{lst:line:stratifiedsamplingForLoop} is an iteration through $N-S$ instead of $N-1$ as in the original HAC algorithm. 
If $S$ is $30\%$ of $N$ for instance, the computation becomes $1-0.3=0.7$ time faster than looping through all $N$. Specifically, the algorithm uses extra memories to store the next-best-merge array NBM to improve the complexity \cite{Manning2008}. The $Cluster()$ function in line \ref{lst:line:stratifiedsamplingCluster} then converts the node linkage table $A_1$ to a tree structure and returns the clusters $C_S$, which stores the instance indices for each cluster. Line \ref{lst:line:stratifiedsamplingSample} implements the sampling algorithm explained above.
We calculate a transformed L2-norm distance between instance $a_{i}^{T}$ and $a_{j}^{T}$ which have NULL entries as
\begin{equation}
dist(x_{i}^{T}, x_{j}^{T}) = \frac{\sqrt{\sum_{t=1}^{T}{({x_{i}^{t}} - x_{j}^{t}})^{2}}}{T}, \text{ if } x_i^t \neq \emptyset \text{ and } x_j^t \neq \emptyset 
\end{equation}
where the numerator is the shortest L2-norm distance between two hyper-planes formed by the non-null dimensions and the denominator penalizes non-null dimensions. 

Using HAC-based stratified sampling benefits the attribution analysis from several properties. First, it guarantees enhanced precision and population depiction compared to other sampling methods. Second, the sampling running time is inversely proportional to sample size $K$, and therefore faster than sampling with other clustering algorithms such as k-means when K is large. compared to k-means, HAC-based sampling methods do not have a convergence problem either. Third, the algorithm is scalable to the visual summary for a large dataset. Additionally, without data prerequisites such as particular distribution assumptions and high dimensional density, the method scales for different applications (\textbf{DG5}). Last but not least, stratified sampling is intimately coherent with the following within-class temporal pattern summarization.

\subsection{Within-Class Summary with NoRCE}
In correspondence with \textbf{DG2}, ViSFA aims to provide sequence summaries for users to recognize temporally contributing patterns in each class. We introduce NoRCE -- a systematical algorithm for noise reduction and the number of cluster estimation. 

A few signature data properties brought challenges to summarizing sequence patterns.
First, RNN training produces significant noise. Eliminating the influence of noise is essential before further operations. 
Besides, high-dimensional data is often sparse. Many algorithms are designed for dense datasets which will produce artifacts for sparse data analysis. We will show an example later.
Additionally, determining the number of clusters $K$ is difficult for unknown data distribution. Although dimension reduction techniques such as t-SNE are often used to visualize data distribution, the visualized distribution is only an approximation.  Visualizing dimension-reduced data can be misleading, especially when the feature dimension is high, due to the dimension curse.
Determining the optimal $K$ for such a dataset is more challenging. 
Because deep neural networks are still ``blackbox,'' it's unclear whether sequences from each class would homogeneously form more than one group. NoRCE is designed to resolve these difficulties.

NoRCE first performs noise reduction. We present this procedure in lines \ref{lst:line:NORCENoiseReductionBegin}-\ref{lst:line:NORCENoiseReductionEnd} in Algorithm \ref{algorithm}. Specifically, $HAC()$ is the HAC calculation function that returns the nodes' linkage table $A_2$ created in the process. 
Figure \ref{fig:gapstatistic} illustrates the concept of noise reduction using a simple 2D example, where the instances are shown in Figure \ref{fig:gapstatistic1}. Instance 17 ($I17$) is the outlier in the dataset that homogeneously forms two clusters on the bottom left and top right, respectively.
During the HAC dendrogram building process, newly formed nodes have lower similarities to nearby nodes than early formed nodes, as shown in Figure \ref{fig:gapstatistic3} where node IDs are in progressive order where nodes with smaller IDs form earlier and vice versa. 
Therefore, the dendrogram iteration vs. distance curve is monotone increasing. As shown in Figure \ref{fig:gapstatistic2}, the distance for the example dataset is monotone increasing during the iteration. 
As shown in Figure \ref{fig:gapstatistic3} left, the distance suddenly increases when $I17$ joins the bottom-up building process of the dendrogram. Meanwhile, the iteration-distance curve exhibits an elbow point where a sudden jump of distance value happens at iteration 36 that cut through $I17$ as shown by the horizontal blue line. This phenomenon is not a coincident, as the outliers are distanced from other nodes. We then leverage this property and use the elbow point as a threshold to detect outliers which are the instances covered by dendrogram layers to the right of elbow point. In Algorithm \ref{algorithm}, the $Distances()$ function on line \ref{lst:line:NORCE_distance} computes the iteration-distance curve from $A_2$. And NoRCE computes the elbow point by smoothing the curve and find the maximum absolute second derivative, as shown on line \ref{lst:line:NORCE_elbow}.
As illustrated in Figure \ref{fig:gapstatistic3} left, the algorithm detects $I17$ as an outlier because it's the single instance whose closest distance to other instances is greater than the elbow point. NoRCE computes the elbow point by smoothing the curve and find the maximum absolute second derivative, as shown on line \ref{lst:line:NORCE_elbow}. Users can also adjust the elbow point interactively, as explained in section \ref{section:interaction}. 

\setlength{\textfloatsep}{10pt}
\begin{algorithm}[!htb]
\caption{Contributing Sequence Pattern Summary \\ for RNN Predictions}
\label{SPS}
{\setstretch{0.55}{\fontsize{7}{7}
\begin{algorithmic}[1]
\Procedure{StratifiedSampling}{$<x_1, ..., x_N>, S$}
\State $Y$, number of samples
\For{$j\gets 1: N$}\label{lst:line:stratifiedsamplingHACBegin}
    \For{$i\gets 1: N$}
    \State $C[j][i].sim \gets Similarity(x_i, x_j)$
    \State $C[j][i].index \gets i$
    \State $I[j] \gets j$\Comment{$I$ tracks active clusters}
    \State $NBM[j] \gets \argmax_{X \in C[j][i]:j \neq i} X.sim$
    \EndFor
\EndFor
\State $A_1 \gets []$
\For {$n \gets 1: N-S$} \label{lst:line:stratifiedsamplingForLoop} 
\State $i_1 \gets \argmax_{i:I[i]=i} NBM[i].sim$
\State $i_2 \gets I[NBM[i_1].index]$
\State $A_1.Append(<i_1, i_2>)$
    \For {$i \gets 1: N$}
    \If {$I[i] == i \wedge i \neq i_1 \wedge i \neq i_2$}
    \State $C[i][i_1].sim \gets max(C[i_1][i].sim, C[i_2][i].sim)$
    \State $C[i_1][i].sim \gets C[i][i_1].sim$
    \EndIf
    \If {$I[i] == i_2$}
    \State $I[i] \gets i_1$
    \EndIf
    \EndFor
\State $NBM[i_1] \gets \argmax_{X \in C[i_1][i]: I[i]=i \wedge i\neq i_1} X.sim$
\EndFor \label{lst:line:stratifiedsamplingHACEnd}
\State $C_S \gets Clusters(A_1)$ \label{lst:line:stratifiedsamplingCluster}
\State $W \gets []$
\For {$s \gets i, S$}
\State $W.Append(RandomSample(C_S[s]))$ \label{lst:line:stratifiedsamplingSample}
\EndFor
\State \Return W
\EndProcedure

\Procedure{NoRCE}{$W, <x_1, ..., x_N>$}\label{lst:line:NORCEbegin}
\Procedure{NoiseReduction}{$W$}\label{lst:line:NORCENoiseReductionBegin}
\label{procedure2}
\State $A_2 \gets HAC(W)$ \Comment{Run HAC on W, return $A_2$}
\State $d \gets Distances(A_2)$\Comment{Converts table $A_2$ to tree} \label{lst:line:NORCE_distance}
\State $ELBOW \gets \argmax_{index}SecondDerivative(Smooth(d(index)))$\label{lst:line:NORCE_elbow}
\State $Y \gets []$ \Comment{leaf nodes before elbow}
\For {$i \gets A_2[1: ELBOW]$ }
    \If{i < S - 1}
    \State $Y.Append(i)$
    \EndIf
\EndFor
\State \Return Y
\EndProcedure\label{lst:line:NORCENoiseReductionEnd}

\Procedure{Adaptive Gap Statistics}{$<x_1, ..., x_N>, Y$}\label{lst:line:NORCEAGSbegin}
\State $N_{ref}$, number of references
\State $K_{max}$, maximum number of clusters
\State $A_3 \gets HAC(Y)$
\State $Gaps \gets []$
\For {$index, K \gets 1 : K_{max}$}
    \State $W_{ks}^{\prime} \gets []$  
    \For {$i = 1: N_{ref}$}
        \State $Y^{\prime} \gets AdaptiveReferencing(X, Y.size)$\label{lst:line:NORCE_AGSreferece}
        \State $A^{\prime} \gets HAC(Y^{\prime})$
        \State $clusterLabels^{\prime} \gets CutTree(A^{\prime}, k)$
        \State $\mu_k^{\prime}, C_k^{\prime} \gets ComputeCenters(Y^{\prime}, clusterLabels^{\prime})$
        \State $W_{ks}^{\prime}[i] \gets \log{\sum_{k=1}^K \sum_{x_i \in C_k^{\prime}} \norm{x_i - \mu_k^{\prime}}^2}$
    \EndFor
    \State $W_{k}^{\prime} \gets \sum_{i=1}^{N_{ref}} W_{ks}^{\prime}$
    \State $clusterLabels \gets CutTree(A_3, k)$
    \State $\mu_k, C_k \gets ComputeCenters(Y, clusterLabels)$
    \State $W_k \gets \sum_{k=1}^K \sum_{x_i \in C_k} \norm{x_i-\mu_k}^2 $
    \State $Gap_k \gets W_{k}^{\prime} - \log{W_k}$
    \State $Gaps.Append(Gap_k)$
\EndFor
\State $OptK \gets \argmax_{k} Gaps[k]$
\State \Return OptK
\EndProcedure \label{lst:line:NORCEAGSend}
\State $OptClusters \gets CutTree(A_3, OptK)$
\EndProcedure
\end{algorithmic}
}}
\label{algorithm}
\end{algorithm}


After removing the outliers, NoRCE introduces an Adaptive Gap Statistic (AGS) algorithm to estimate the number of clusters. Gap Statistic (GS) \cite{Tibshirani00} is a method for determining the number of clusters in a set of data by comparing the within-cluster dispersion to its expectation under an appropriate reference distribution of data. The GS method is proved to outperforms other numbers of cluster estimating methods, such as the average silhouettes method \cite{Rousseeuw1987} and the elbow methods \cite{KETCHEN1996}. Figure \ref{fig:gapstatistic4} shows the results of the gap statistic before (red) and after (black) removing the outlier. The estimated cluster numbers are correspondence with the most significant gap values (yellow). After removing the outlier, the algorithm automatically detects two optimal clusters with the cut that splits the largest gap in the dendrogram, as shown in Figure \ref{fig:gapstatistic4} right.

We show the AGS algorithm in lines \ref{lst:line:NORCEAGSbegin}-\ref{lst:line:NORCEAGSend} in Algorithm \ref{algorithm}. AGS takes two parameters: the number of references $N_{ref}$ and the maximum cluster number $K_{max}$, and returns the estimated number of cluster $OptK$. We set $N_{ref} = 3$ and $K_{max} = 10$ in our experiments. The classic gap statistic algorithm creates randomized data reference that is within the original data range. However, for high-dimensional sparse data, such references often greatly change the original data distribution and lead to unreasonable cluster number estimation. Imagine a dataset where all instances are individual points in a 3D space. Sparse data indicates that most points in this space are on the planes made by two axes or on one of the axes. However, randomly generated data would evenly distribute in the 3D space, which is different from the sparse data distribution. Then the question is how to reference data that has similar distribution? Fortunately, NoRCE's input $Y$ is sampled from the original training data $X$. As shown in line \ref{lst:line:NORCE_AGSreferece} in Algorithm \ref{algorithm}, function $AdaptiveReferencing()$ randomly samples the same number of instances as $Y$ from the residual data $X-Y$ as references. Then computes the gap values through the for loops and returns the optimal cluster number $OptK$, which correspondences with the largest gap value. Finally, line \ref{lst:line:NORCEAGSend} computes the estimated clusters for later visualization. Specifically, users can also adjust the number of clusters interactively. We introduced more details in the next section.


\begin{figure*}
 \centering
 \includegraphics[width=\textwidth]{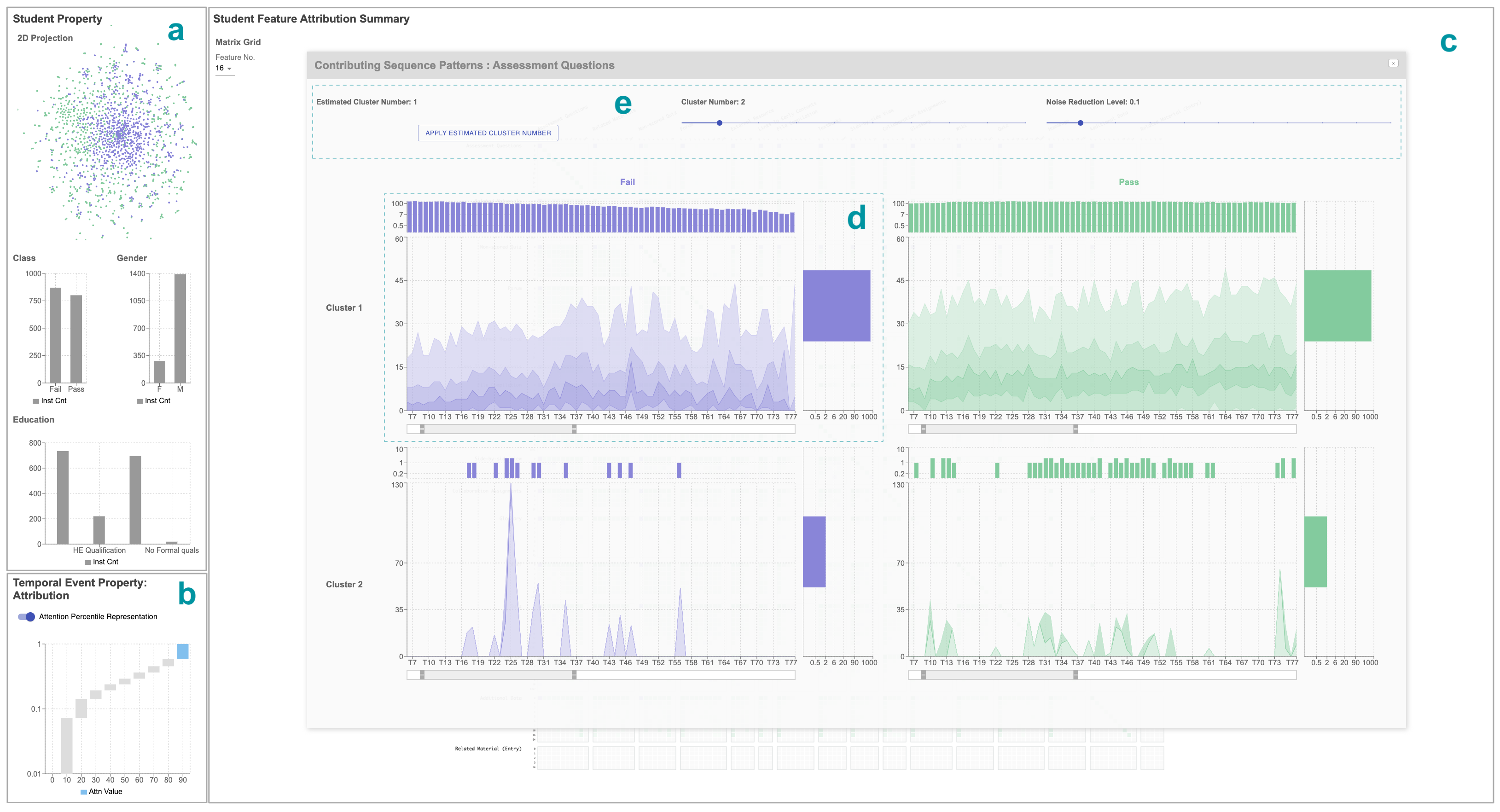}
 \caption{The interface of ViSFA is composed of a) a instance property view including a instance 2D projection view and an attribute view, b) a feature attribution chart and c) a summary view, which is initialized as a matrix grid view that visualizes the attribution of feature values. Clicking a matrix in the matrix grid triggers the temporal pattern view, as shown in the ``contributing sequence patterns'' window.}
 \vspace{-10px}
 \label{fig:visfa}
\end{figure*}

\subsection{Visualization Design}
   
In this subsection, we focus on the visualization of contributing temporal pattern summaries and their comparison between prediction classes, based on the algorithm presented in the previous subsection.


\subsubsection{Dashboard}
We divide the dashboard on the left of Figure~\ref{fig:visfa} into two parts. 
We visualize the statistical distributions of instance-level attributes on the top (a) and the event-level attention distribution on the bottom (b).

To provide users with an overview, the projection view embeds the high-dimensional temporal feature sequences into a two-dimensional canvas via t-SNE~\cite{maaten2008visualizing} (see Figure~\ref{fig:visfa}). The dimension reduction result can reflect the similarity of instances in a birds-eye view. 
Besides, the distributions of instance attributes are indispensable for users to review the data in partitions. For example, student e-learning behavior analysts always want to understand the distributions of students’ gender and education level.
ViSFA lists bar charts in the attribute view (see Figure~\ref{fig:visfa}b for all instance attributes. Specially, we employ a contrast color pair, purple and green, to encode the bars for two classes. This color scheme is consistent in the entire ViSFA system.

At the temporal event level, the old version of ViSFA uses a slider bar to let users filter events by their contribution. Based on users’ feedback, the improved ViSFA provides the histogram of normalized contribution, where the horizontal axis represents the contribution ranges from 0 to 1 with a 0.1 step. Because the distributions of attribution scores can vary largely for different RNN models, ViSFA provides another mode to easily filter top-contributing events. In this mode, the x-axis represents the percentiles, and the y-axis represents the corresponding ranges of the attribution score, as shown in Figure~\ref{fig:visfa}(b). Users click the switch on the top of this view to switch these two modes. The histogram shows the contribution distribution of temporal events, while the percentile mode illustrates what attention range is covered by each percentile.

\subsubsection{Temporal Pattern View}

ViSFA illustrates the over-time feature value change in the temporal pattern view, as shown in Figure~\ref{fig:visfa}c.
AttentionHeatmap~\cite{WangONM18} overlays contributing temporal events in a time-value plane. Such design has perception issues, such as line crossing hinders users to notice jump over edges. In this work, the temporal pattern view visualizes the NoRCE clustering results using stacked area charts. 

As mentioned earlier, ViSFA provides noise reduction for the analysis using the elbow method. In the temporal pattern view, ViSFA should provide an interface for users to remove noise based on the elbow method. Dendrograms in the elbow method (Figure~\ref{fig:gapstatistic}b) illustrate the clustering loss of each iteration by distance on the vertical axis. However, it is unnecessary for users to distinguish each individual by ID on the horizontal axis. Besides, we design ViSFA for domain analysts who may not have a computer science background. Thus, we simplify the view and use a slider bar to let users select different noise reduction levels, as shown in Figure~\ref{fig:visfa}c3. We compute the noise reduction level as the number of instances at the elbow point over the total number of instances.

We visually summarize the denoised instances from two classes in a juxtaposition layout for easy comparison~\cite{gleicher2011visual}. 
We sort the clusters belonging to each class in a size descending order from top to bottom. In each cluster, the temporal pattern summary (Figure~\ref{fig:visfa}d) is composed of a horizontal bar chart on the right, an area chart on the bottom-left, and a bar chart on the top. 

The horizontal bar chart shows the cluster size. For easy comparison of cluster sizes in different ranges from all clusters, the horizontal axis is log-scaled.  We design the area chart to visually summarize the over-time changes in the feature values across all the instances in the cluster. The design of the area chart is inspired by boxplots that statistically depict the quantiles of a group of values. The horizontal axis represents time and the vertical axis represents the value of the selected feature. The top edge of each area connects the same quantiles horizontally for the ability to handle a large number of time steps compared to the boxplot. And the area chart is also more flexible in increasing the number of quantiles.
Five area layers represent a set of predefined quantiles in Figure~\ref{fig:visfa}. The middle quantile is colored with the darkest intensity and the intensities decrease while the quantiles go further away from the middle (towards top and bottom). 
The bar chart on the top is a contribution indicator that shares the horizontal axis with the area chart. The log-scaled vertical axis denotes the number of instances contributing to the pattern in the bottom area chart.

To demonstrate temporal patterns involving two features, AttentionHeatmap~\cite{WangONM18} applies two-hierarchy axis, which is not intuitive to read values or compare trends over time. ViSFA improves the design by placing the axes for the two features symmetrically, as shown in Figure 1
in the supplementary material. It is convenient to discover relationships from a vertically symmetrical design. 


\subsection{Interactions}
\label{section:interaction}

ViSFA provides a variety of interactions for users to explore data and discover insights from both instance level and temporal event level. 
For instance exploration, users can interact with the t-SNE projection view by lasso-selecting a group instance to review their attribute distribution. Users can also click any bar(s) in the bar charts to filter the instances with the corresponding value. For instance, selecting the high qualifications the “Education” bar chart in Figure~\ref{fig:visfa}(a) results in changes in other views (e.g. a better pass/fail rate). 

For the exploration of temporal events, users can:

\textbf{Filter by attention.} The neural attention models calculate the attention of events based on their contribution to the prediction. Taking this advantage, ViSFA lets the users filter the events by their contribution. As mentioned earlier, ViSFA provides two modes for filtering the temporal events by their contribution value. In both modes, each bar is associated with a contribution range.  Users can click the bar to use the corresponding contribution range to filter the temporal events in both classes. Multiple selections are enabled and clicking a bar again will deselect the bar. The filtering interaction results in an update in the matrix grid view and the temporal pattern view. The filter range can be set by users via a slide bar.

\textbf{Check temporal patterns.} The temporal pattern view can be triggered by clicking a cell in the matrix grid view. The cells on the diagonal correspond to temporal patterns with a single feature, and the other cells correspond to temporal patterns with two features. 
The temporal pattern view provides multiple interactions for users to easily compare temporal patterns in different classes/clusters. 
Users can hover any component in the temporal pattern summary and all corresponding components in all clusters and classes will respond. For example, hovering any horizontal bar chart enables the comparison between the cluster sizes among all clusters. And hovering any time-step in the area chart triggers the tooltips in all clusters, each of which shows the feature values of all percentiles at the hovering time-step. 

On the bottom of Figure~\ref{fig:visfa}d, we design a slider bar for the interaction. Users can adjust either side and the position of the slider bar to locate a temporal focus. The position of the slider bar relative to the total slider length indicates the selected temporal range relative to all time-steps. 

\textbf{Use concisely designed UI to operate complex background algorithms.}
ViSFA provides a few widgets in the interaction panel (Figure~\ref{fig:visfa}e). 
\textbf{Noise reduction.} It is challenging to decide which records are noise for various datasets by automatic models. Different decisions may lead to distinct results in the latter analysis~\cite{quinlan1986induction}.  Thus, we allow users to set the parameter for noise different levels of noise reduction.  
In an earlier design, we let the users adjust the noise reduction level by dragging a dashed line in the line chart illustrating HAC. Based on the user feedback, a slider bar indicating the level of noise reduction is preferable, as shown in Figure~\ref{fig:visfa} right. The interaction panel also provides the users with a button to apply the automatically \textbf{estimated number of cluster}. 
Meanwhile, the model-defined cluster number may be unsatisfying for distinct application scenarios~\cite{kim2004cluster}. Users may want to explore clusters freely, so the interaction panel enables a slider bar in the middle for adjusting different numbers of clusters.





\section{Experiments and Results}

We verified the effectiveness of our approach with two open datasets. We trained two RNN attention models: LSTM and bidirectional LSTM (BiLSTM) for the analysis of both datasets. All models are cross-validated. The first dataset is from the education domain for studying students' learning activities. The second dataset is from the medical domain for studying patients mortality study. In the following subsections, we showcase the visual analytics results that help understand RNN models’ behavior. The discovered insights also guide future decision-making towards achieving desired outcomes.

\subsection{Open University Learning Analytics (OULAD)}

In this use case, we validate ViSFA with the OULAD dataset~\cite{oulad}.  We train RNN models to predict online course evaluation results: either \textcolor{mypurple}{fail} or \textcolor{mygreen}{pass}. The LSTM model makes predictions at a 75\% accuracy, and the BiLSTM 88\%. 180 days of more than 15 hundred students’ interaction records are used - each day as a temporal event. Each temporal event contains 16 features.  Each feature is a webpage in the online course system, and the feature values are the number of visits in one day. The webpages are such as homepage, course contents, assignments, quiz, glossary, forum, supplementary contents, etc.

The feature contribution ranking results from interpolating two RNN models indicate that ``assessment questions, '' ``non-scored quiz,'' and ``forum''  have the most significant contributions among all webpages. Contrarily, pages such as ``glossary'' and ``additional data'' are the least contributing pages. 
As one might expect, the assessments and quizzes are important in examining students’ learning achievement. The amount of a student’s interaction with the related webpages is expected to be significantly correlated with the course’s final performance. 
In addition, high participation in the forum is effective in achieving satisfying course evaluations. Studies show that peer learning can enhance learning outcomes such as increased motivation and engagement in the learning task, deeper levels of understanding, increased metacognition, the development of higher-order thinking skills and divergent thinking \cite{Blumenfeld1996, Thomas2002}. Meanwhile, the pages like ``glossary'' and ``additional data'' are obviously less related to a course evaluation and thus rank lower.

\begin{figure}[hbt]
    \centering
    \subfloat[\label{fig:oulad-raw} Raw data]{%
         \centering
         \includegraphics[width=\columnwidth]{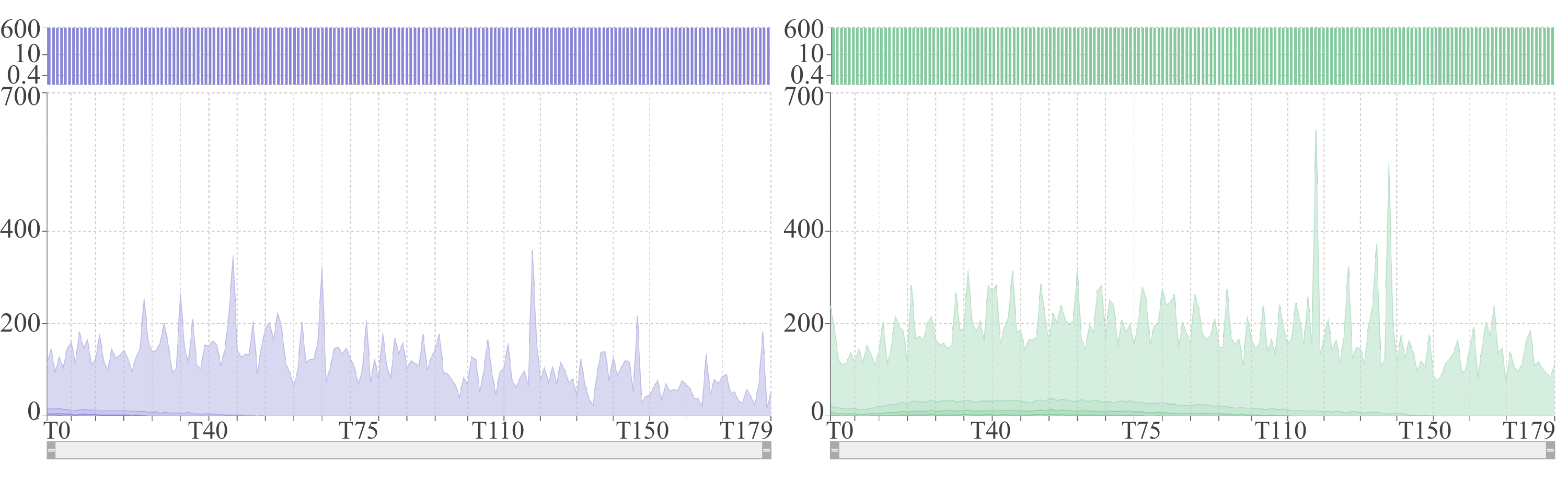}
    }
    \vfill
    \vspace{-10px}
    \subfloat[\label{fig:oulad-lstm} Top 10\% contributing events: LSTM] {%
        \centering
        \includegraphics[width=\columnwidth]{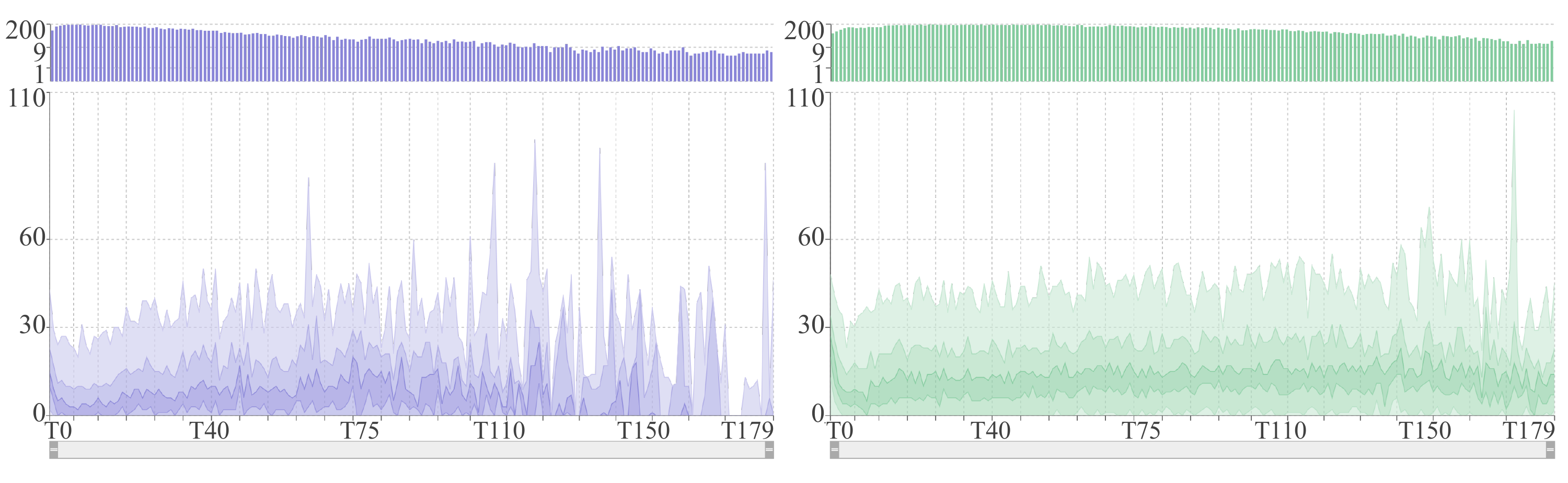}
    }  
    \vfill
    \vspace{-10px}
     \subfloat[\label{fig:oulad-lstm-norce} Top 10\% contributing events: LSTM + NoRCE] {%
        \centering
        \includegraphics[width=\columnwidth]{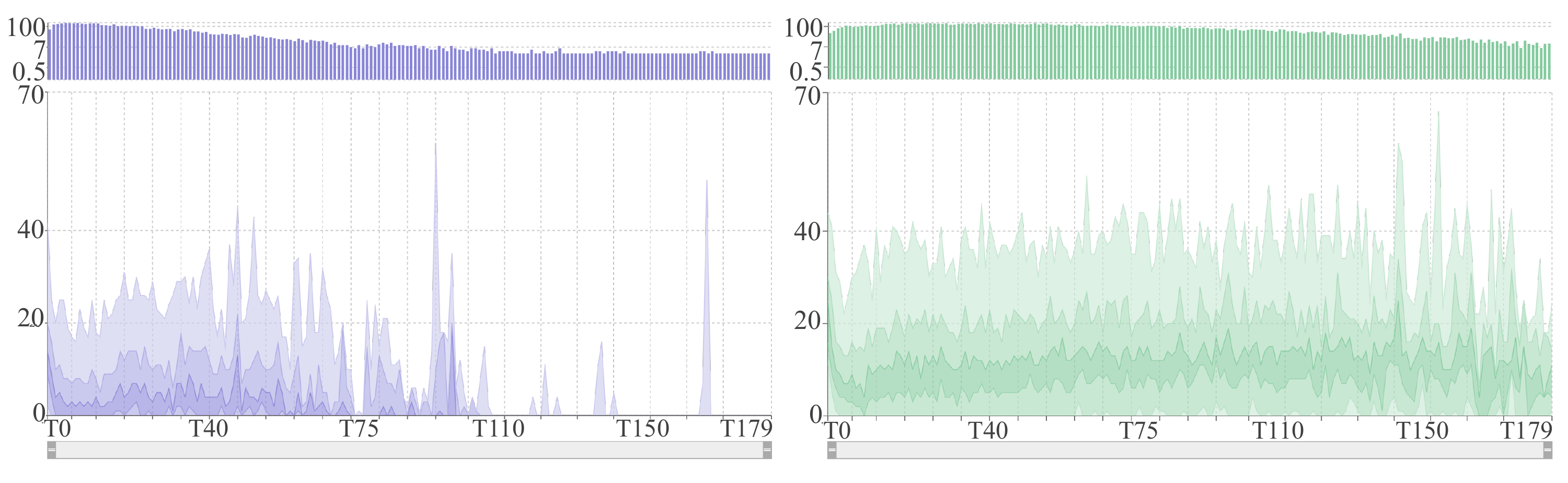}
    }  
    \vfill
    \vspace{-10px}
    \subfloat[\label{fig:oulad-bilstm} Top 10\% contributing events: BiLSTM]{%
        \centering
        \includegraphics[width=\columnwidth]{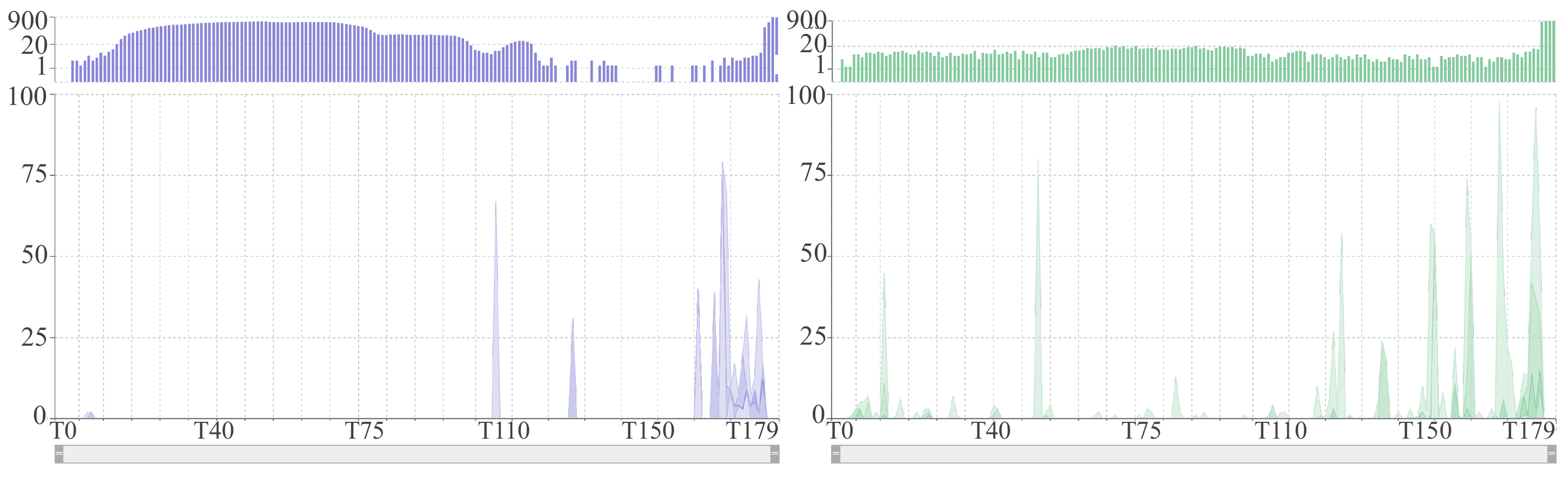}
    }  
    \vfill
    \vspace{-10px}
    \subfloat[\label{fig:oulad-bilstm-norce} Top 10\% contributing events: BiLSTM + NoRCE]{%
        \centering
        \includegraphics[width=\columnwidth]{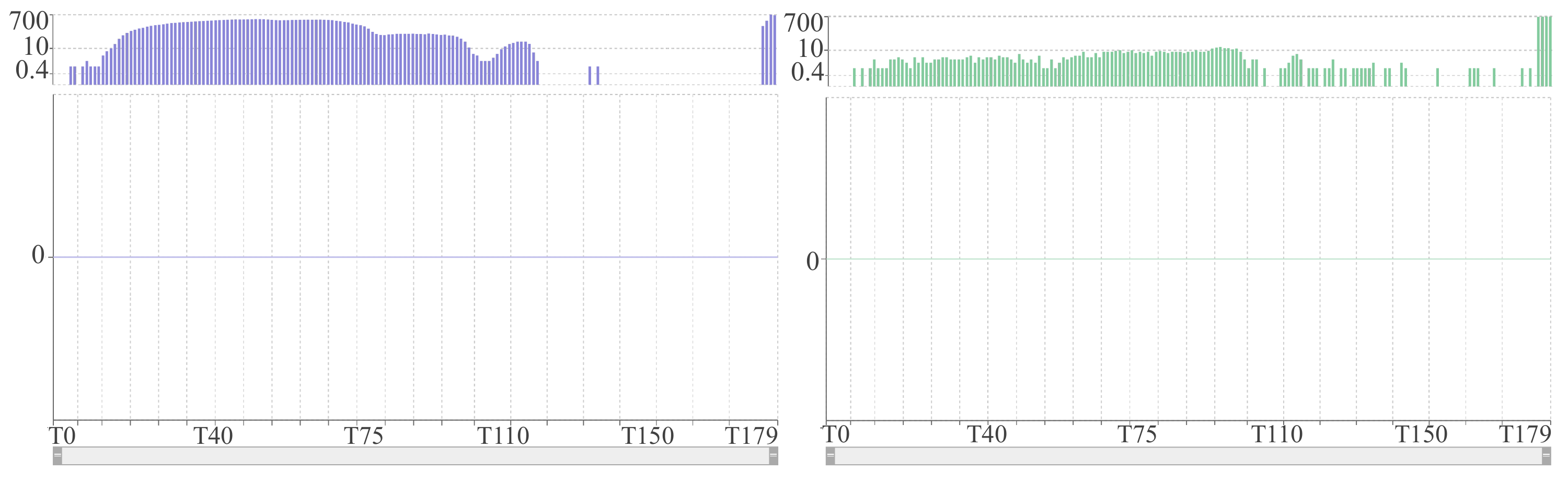}
    }  
    \vspace{-10px}
    \caption{Temporal pattern summaries for feature ``assessment questions.''}
 \label{fig:oulad-assessment}
\end{figure}

Knowing what features have the highest contribution is meaningful in tasks such as feature selections, but less critical in guiding future decision-making. 
The temporal pattern summary help to discover contributing temporal patterns and the contribution bar chart on the top quantifies the contribution for each time-step. 
We pick the feature ``assessment questions'' as an example to showcase the temporal pattern analysis results. Figure~\ref{fig:oulad-assessment} illustrates the effectiveness of the designed algorithm and visualization in revealing the underneath contributing temporal pattern using two RNN models.


Figure~\ref{fig:oulad-raw} shows the temporal pattern summary for the raw data. The real-world data is sparse and noisy. We can hardly see any patterns within either class due to the data sparsity, where a lot of zeros ``occupy'' many quantiles that only the top quantile is shown. Theoretically, if a group of students has similar learning behavior over time, the area chart is expected to show similar fluctuations across all percentiles. Therefore, Figure~\ref{fig:oulad-raw} shows no similar patterns among the students if we look at the raw data.
In addition, the data noise introduces a lot of spikes on the top edge of the area chart. Because raw data covers all events, the contribution indicators simply show the number of all instances for all time steps.

The visual summaries of temporal patterns learned by two the LSTM and BiLSTM models are shown in bottom four sub-figures in  Figure~\ref{fig:oulad-assessment}. The visualization results based on two models both indicate a single dominant cluster, where other clusters only have one or two instances if increasing the number of clusters manually. Consistently, the number of cluster estimation output ``one'' for both models. This indicates that the RNN models recognize a mutual pattern from all instances for both \textcolor{mypurple}{fail} and \textcolor{mygreen}{pass} classes. 

Figure~\ref{fig:oulad-lstm} and Figure~\ref{fig:oulad-bilstm} show the visual summaries of the temporal events that have the top 10 percent contribution to the prediction. Figure~\ref{fig:oulad-lstm-norce} and Figure~\ref{fig:oulad-bilstm-norce} are the final results produced by the NoRCE algorithm. 
Comparing the results before and after applying NoRCE, we can see that the NoRCE algorithm can help clean noisy results produced by the RNN attention models. Specifically, the noise reduction mainly removes the events of low contributions (short bars in the bar charts.) The remaining temporal patterns and contribution patterns are more smooth for users to discover insights.

Figure~\ref{fig:oulad-lstm-norce} shows a decreasing contribution over time for both classes, which means the temporal sequence patterns in the early time steps have higher contributions. Both area charts show relatively consistent over-time fluctuations among the percentiles. As demonstrated in the video, a user further explores more details of this result. She adjusts the temporal focus to the first month, and slide the temporal focus rightwards to review the patterns in the next three months. The results in the \textcolor{mygreen}{pass} class show no significant change in the values for all percentiles. Contrarily, the value of the \textcolor{mypurple}{fail} class decreases significantly after the second month. These signals indicate that a student’s interaction with the assessment questions in the first three months has a higher contribution to their course evaluation results. And, a persistent interaction is important in achieving the desired course performance.

Figure~\ref{fig:oulad-bilstm-norce} shows temporal pattern summary for the BiLSTM model. The results in the area chart and the contribution indicator shows the BiLSTM model successfully captures more common patterns among the students. Especially for the \textcolor{mypurple}{fail} class, the contribution indicators show the number of instances making the contribution is close to the total number of instances. And the pattern in the area chart is all zeros on all time-steps, which indicates that failing to interact with assessment questions from day 30 to day 80 is convincingly correlated with the bad course performance.
Interestingly, the results learned by BiLSTM are more contrastive between two classes. The \textcolor{mypurple}{fail} class shows an extremely certain pattern due to the high contributions for a long time period, while the \textcolor{mygreen}{pass} class does not show highly contributing temporal patterns as the contributions in most time-steps are less than ten.


\begin{figure}
 \centering
 \includegraphics[width=\columnwidth]{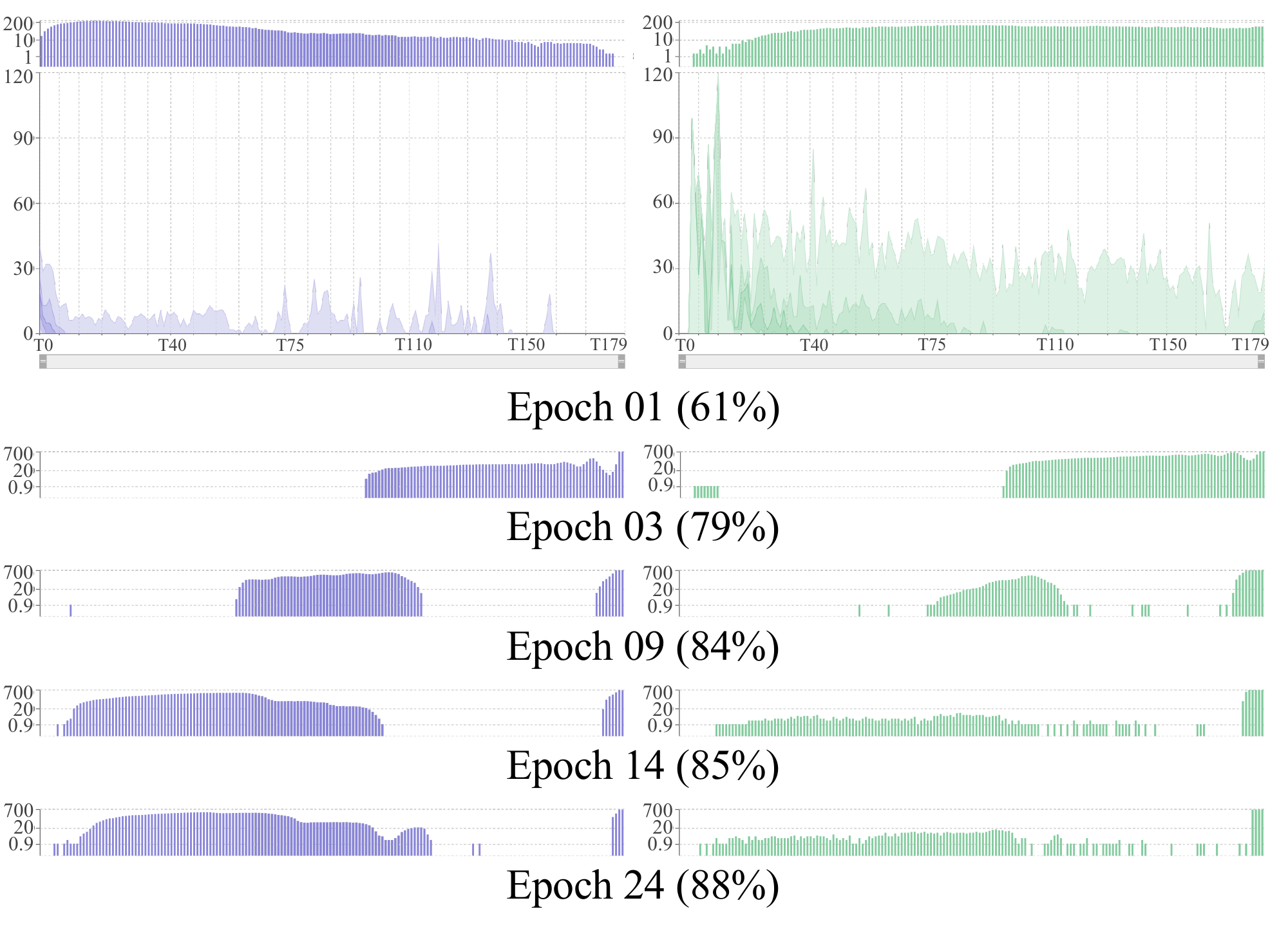}
 \caption{Graduate changes during the training training process, from the epoch 01 to epoch 24 (best performance).}
 \label{fig:biLSTM-epochs}
 \vspace{-5px}
\end{figure}

\textbf{Model behavior in the learning process.}
Figure~\ref{fig:biLSTM-epochs} shows the temporal summary changes of the BiLSTM model, with their accuracy shown in parentheses. Same as Figure~\ref{fig:oulad-bilstm-norce}, the area charts from epoch 03 to epoch 24 show zeros for all time-steps and thus omitted. We can notice the model is gradually searching important time-steps around all time-steps during the learning process for both classes. 
For the \textcolor{mypurple}{fail} class on the left column, the top edge of the bars in the contribution indicators all connect to smooth line shapes. The contribution indicator at the first epoch shows a more evenly distributed pattern over time, as the learning just started. The contribution indicators become more concentrated in certain time periods for the later process.
The \textcolor{mygreen}{pass} class on the right column also shows an evenly distributed pattern at the first epoch. At epoch 03 and 09, the \textcolor{mygreen}{pass} class show very similar patterns to the \textcolor{mypurple}{fail} class. However, the pattern in the \textcolor{mygreen}{pass} class becomes very different from the \textcolor{mypurple}{fail} class while the learning continues in the later stages.
This illustrates the model captures more common patterns from the instance in the \textcolor{mypurple}{fail} class to distinguish two classes at the final step.

\subsection{EHR Data}
The development of ViSFA is first motivated by the needs of ICU patients' mortality analysis. 
We validate ViSFA with the MIMIC \cite{mimiciii} dataset. We train the RNN models using four categories of patients' medical records -- ICU stay, inputs, procedure and lab results. The records contain 37 temporal features, as shown Table 1 in the supplemental material. The models use patients' records in the first 48 hours after admission to predict their mortalities after 48 hours. The analysis task is to find what features and values among all features contribute most to mortality predictions and what is the difference of temporal patterns in contributing feature values between mortality prediction classes.
The preprocessed dataset contains 14K patients (instances) including 1.2 million temporal events, where the alive/dead patient ratio is 10:1. We train RNN models using the oversampling technique so the interpretation is based on balanced data. Then 10\% data are sampled in the stratified sampling procedure. 

The visualization results illustrate that most top-ranked features are the ``inputs'' -- any fluids which have been administered to the patients. The results indicate ``inputs'' are more important in keep patients alive compared to ``ICU events,'' ``procedures,'' and ``lab tests.'' According to the domain scientist, the result is reasonable as fluids affect the cardiovascular, renal, gastrointestinal, and immune systems in critical illness treatment \cite{Martin2018}. 

The features of the highest contribution are ``drips,’’ ``antibiotics (Non IV),'' and ``prophylaxis.’’
Take the ``antibiotics (Non IV)'' as an example, the raw data does not show any common pattern among the data quantiles. However, the contributing temporal pattern summary in ViSFA (Figure 2 in the supplemental material) for ``Antibiotics (Non IV)'' shows contrastive temporal patterns for the ``dead'' and ``alive’’ classes. 
From the visualization results, we notice several phenomenons. First, the contributing temporal events for the ``dead'' class show higher antibiotic levels throughout the entire time period comparing to the alive class for the LSTM model. The BiLSTM model shows a consistent result except that the BiLSTM suggests only the last two time-steps have an extremely large contribution.
According to the domain scientist, combining antibiotics is a strategy often used by clinicians, but recent research shows that such a strategy can cause body resistance \cite{Wright2016}. Also, there is research demonstrating that deploying synergistic antibiotics can, in practice, be the worst strategy if bacterial clearance is not achieved after the first treatment phase \cite{Pena-Miller2013}, which can be the explanation of the phenomenon.
Second, the antibiotic dosage for two classes both fluctuated throughout time, but their temporal patterns show significant differences. 
For instance, the alive class feature in a relatively more steady fluctuation compared to the dead class, which shows more significant peaks and bumps. Specifically, the first great drop of value happens at 4h for the alive class but happens at 8h for the dead class. Both classes have fluctuated patterns around 24h ($\pm2$h). However, the alive class shows a more steady change afterward while the dead class shows two significant peaks in the next six hours, and the dosage becomes steady afterward. 
The domain scientist who participated in the case study indicates that she is highly interested in the visualized result. She considers the comparative temporal pattern visualization is informative in helping to understand what evidences are used by the AI to make predictions. Although she can not verify the cause of antibiotics dosage change on every time point without further experiments, she states the visual summaries could potentially reveal critical time points in the patient treatment process.
Besides, she verifies that the suddenly increased values in the last two hours for the dead group are highly reasonable because the signal indicates previous antibiotics were not effective enough.


\section{Discussion}
During the exploration of feature attribution for two fundamentally distinctive datasets, we found a mutual phenomenon that there is always a \textbf{dominant concentrate cluster}. Enlarging cluster number hardly changes the dominantion, and only small clusters appear in the bottom, as shown in Figure~\ref{fig:visfa}. The NoRCE computation results also verify this phenomenon that the estimated cluster number is one in most experiments. 
We consider the mutual temporal patterns in the dominant cluster as the important facts that contribute to distinguishing two prediction classes. These facts can be complex, as illustrated in the EHR data analysis where domain expert suggests more longitudinal experiments are required to verify its reliability. And the facts can be straightforward to understand and explain, as shown in the OULAD data analysis. However, we can not conclude that RNN models always learn one dominant pattern for each feature without abundant experiments in various application domains. Therefore, we design the temporal view so that the analysis is feasible for visualizing more large clusters.

\textbf{Closed-loop training.}
In the OULAD example, we initially trained an LSTM model using all 285 days. The contribution visualization shows that only the first 180 days have a high contribution to distinguishing two prediction classes. We then retrained the model with only 180 days of the data and the approach improves the accuracy by 4 percent. This demonstrates the visualization results derived from ViSFA can help develop a closed-loop RNN model training.

\textbf{Causality vs correlation}. Although our approach can visualize features and their overtime patterns contributing to class predictions, it's inadequate to consider top-ranked features and their overtime patterns the cause to the predicted class. The learned attribution reflects what features/values in the training dataset significantly correlated to the prediction whereas causation indicates that the labeled class is the result of the features, which is a stronger statement to prove and beyond the scope of our discussion.
However, although it's insufficient for causal analysis, feature attribution analysis provides evidence for further causal analysis. 
For example, students' behavior learned by RNN can not be proved to cause their course evaluation results without further controlled experiments~\cite{Pearl2009}. But our approach uncovers important pages and students' interactive temporal patterns that are correlated with the course evaluation results. 
Similarly, the antibiotics (Non-IV) doses applied to the patients who died after 48 hours are not necessarily the cause of their death. But the analysis provides statistically supportive evidence that different doses of antibiotics (Non-IV) are correlated with patients mortality.

The ViSFA system uses a \textbf{modular design} where the stratified sampling module is for scalable and comparative analysis, and the NoRCE module is for noise reduction and temporal pattern summarization. ViSFA also provides a systematic framework containing these modules to motivate more discussions to probe RNN classifiers’ behavior in the value-level.
As discussed in the related work section, the proposed analysis can be used to a broad range of analysis scenarios and the noise removal and the pattern summarization modules are two requisite steps in the analysis. 


\section{Conclusion}
As deep learning pervasively used in decision-making tasks for multidimensional sequential data analysis, it's essential to understand contributing features and temporal patterns for predictions. In this work, we present ViSFA, the first visual analytics system that scalably summarizes value-level feature attributions with recurrent neural attention networks. We test ViSFA with two real-world datasets, each using two RNN models. The case study results demonstrate that ViSFA can 1) help distill contributing patterns for different RNN models of different prediction performances, 2) reveal gradual changes in the RNN model learning process, and 3) help effectively reason value-level feature attribution for different application domains, and the visual summaries of temporal patterns in feature attribution provide guidelines for making future decisions. We hope our work will motivate further research in developing domain-user-oriented analysis systems with deep learning.


\bibliographystyle{abbrv-doi}

\bibliography{visfa}
\end{document}